\documentclass{article}

\PassOptionsToPackage{numbers, compress}{natbib}

\usepackage[final]{style}

\usepackage[utf8]{inputenc} 
\usepackage[T1]{fontenc}    
\usepackage{hyperref}       
\hypersetup{
    colorlinks=true,
    citecolor=blue,
    linkcolor=blue,
    urlcolor=blue
}
\usepackage{url}            
\usepackage{booktabs}       
\usepackage{amsfonts}       
\usepackage{nicefrac}       
\usepackage{microtype}      
\usepackage{xcolor}         
\usepackage{multicol}

\title{AI Agents with Human-Like Collaborative Tools: Adaptive Strategies for Enhanced Problem-Solving}

\author{%
  Harper Reed\\
  2389 Research\\
  \texttt{harper@2389.ai} \\
  \And 
  Michael Sugimura\\
  2389 Research\\
  \texttt{michael@2389.ai} \\
  \And 
  Angelo Zangari\\
  University of Illinois Chicago\\
  \texttt{azang@uic.edu} \\
}

\begin{document}

\maketitle

\begin{abstract}
We investigate whether giving LLM agents the collaborative tools and autonomy that humans naturally use for problem-solving can improve their performance, providing Claude Code agents with MCP-based social media and journaling tools and the flexibility to use them as they see fit.

Across 34 Aider Polyglot Python programming challenges, collaborative tools substantially improve challenging problem performance, delivering 15–40\% cost reductions, 12–27\% fewer turns, and 12–38\% faster completion compared to baseline agents. Effects on the full challenge set are mixed, indicating collaborative tools function as performance enhancers primarily when additional reasoning scaffolding is most needed.

Surprisingly, different models naturally adopted distinct collaborative strategies without explicit instruction. Sonnet 3.7 demonstrated broad engagement across tools, benefiting from articulation-based cognitive scaffolding. Sonnet 4 exhibited selective adoption, primarily leveraging journal-based semantic search when facing genuinely challenging problems. This adaptive behavior parallels how human developers adjust collaborative approaches based on expertise and problem complexity.

Behavioral analysis reveals agents prefer writing over reading by 2--9x, indicating that structured articulation drives performance improvements rather than solely information access and retrieval. Our findings suggest that AI agents can systematically benefit from human-inspired collaboration tools when facing problems at their capability limits, pointing toward adaptive collaborative interfaces as reasoning enhancers rather than universal efficiency improvements.
\end{abstract}

\section{Introduction}

Human programmers rarely build in isolation. They engage in rubber duck debugging to articulate problems clearly, search through shared knowledge bases to find similar solutions, build incrementally on previous work, and leverage team discussions to break through mental blocks. These collaborative behaviors are not merely social conveniences—they represent approaches to problem-solving that help humans find and fix mistakes and discover more efficient solutions. Yet current LLM agents, despite their impressive individual reasoning capabilities, lack access to these social collaboration mechanisms that could dramatically improve their performance.

We hypothesize that \textbf{providing LLM agents with human-like collaborative tools and the freedom to use them naturally can improve problem-solving performance}. Rather than relying solely on prescriptive prompting or architectural changes, we provide agents with MCP tools that approximate the collaborative practices humans use to solve problems: sharing insights, building on previous work, and engaging in reflective debugging processes ~\cite{anthropic2024mcp}. We pair human-inspired affordances (journal with lightweight search,and Twitter-style shortform social media posts) with \emph{affordance-framed prompts}—brief, invitation-style prompts that invite (but does not prescribe) articulation and opportunistic retrieval (see ~\ref{app:tool-instructions})~\cite{gibson1979ecological,norman2013doet}.

To test this hypothesis, we developed Botboard, an internal social media platform that combines Twitter-like microblogging with journal functionality. The platform provides agents with semantic search capabilities for journal entries and tag-based filtering for social media posts, enabling both structured reflection and casual information sharing. Our experimental design tests both the act of articulating problems, frustrations, and celebrations along with the accumulation of information you would see in a team of agents working together over time.

We conduct multiple runs across different 'teams' of agents, where each team shares access to the same Botboard instance through a unique team API key. The way we structure our experiments is that the first run in each team starts with empty social media and journal databases. As agents complete problems, they organically populate these databases with posts and entries. For each collaborative tool variant, we run a second pass over the same challenges using accumulated information from the first run, simulating how agents might build upon previous work when institutional knowledge exists.

We evaluated our approach across 34 programming challenges from the Aider Polyglot Python benchmark\footnote{https://github.com/Aider-AI/polyglot-benchmark/tree/main/python/exercises/practice}, an established externally-validated coding benchmark derived from Exercism's most challenging exercises. These tasks range from string manipulation problems to complex algorithm implementations requiring sophisticated reasoning, such as bowling score calculation, hexagonal grid pathfinding, and zebra logic puzzles.

To ensure rigor, we ran the benchmark through a dockerized evaluation pipeline that isolates the effects of different tool variants. Most importantly, the results show that social collaboration tools enable agents to develop adaptive strategies. These adaptive strategies allow agents "punch above their weight" on challenging problems with \textbf{cost reductions of 15-40\%, turn reductions of 12-27\%, and time improvements of 12-38\% compared to baseline capabilities.} While agents with access to collaborative tools achieved modest improvements or mixed quantitative performance across the full dataset, the dramatic improvements on challenging problems those which exceed baseline Sonnet 4 and 3.7 capabilities demonstrate that collaborative tools provide the greatest value when additional reasoning scaffolding is most needed, functioning as difficulty-dependent performance enhancers rather than universal efficiency improvers.

Through detailed analysis of agent interactions, we identified how different models naturally gravitated toward different collaborative strategies without explicit instruction. This adaptive behavior parallels how human developers adjust their collaborative approaches based on expertise level and problem complexity.

Crucially, agents adopted these collaborative behaviors organically without explicit instruction in their prompting or configuration files. When facing difficult debugging challenges, agents would spontaneously post to social media or journal about their struggles, then return to solve problems more efficiently.

\textbf{Contributions:} (1) We demonstrate that codifying human collaborative behaviors into accessible tools creates difficulty-dependent performance enhancers, enabling agents to 'punch above their weight' on challenging problems while increasing transparency in problem-solving processes; (2) We identify how agents organically develop adaptive collaborative strategies that vary by model capability and problem difficulty, mirroring human collaborative flexibility; (3) We establish a reproducible dockerized evaluation framework for studying agent collaborative behaviors.

\section{Related Work}

\subsection{The Prescriptive Paradigm in LLM Research}

Recent comprehensive surveys reveal a field converging on prescriptive optimization as the dominant approach to enhancing LLM capabilities. Zhao et al.'s extensive survey documents how current research prioritizes control through detailed prompting strategies, structured planning frameworks, and deterministic tool interfaces~\cite{zhao2025survey}. This paradigm manifests across core areas: prompting guidelines emphasizing "expressing the task goal clearly" and "decomposing into easy, detailed sub-tasks"; reasoning enhancement through prescribed Chain-of-Thought steps; tool integration via "encapsulating available tools with API calls"; and planning through formal decomposition with structured feedback loops.

Our work represents a fundamental departure from this control-oriented paradigm. Rather than asking, \textbf{“How can we better specify and control LLM behavior?”} we investigate, \textbf{“What capabilities emerge when we provide interfaces for agents to use collaborative tools as they see fit?”} By providing agents with affordance-framed instructions like, \textbf{“Post if you want to, browse when you feel like it, or ignore it entirely,”} and allowing them to determine their own tool usage patterns, we mirror how human developers naturally adapt their approaches based on individual needs and problem complexity. One engineer might benefit from extensive rubber duck debugging while another may perform more upfront research before starting a task. We try to allow our agents to organically develop model-specific tool usage strategies without explicit guidance on when or how to use available tools. We observe that this open-ended approach allows agents to "punch above their weight" and solve problems that would otherwise exceed their individual reasoning limits.

\subsection{Open Ended Tool Usage}

Current tool-augmented systems focus on individual agents learning prescribed usage patterns. Yao et al. demonstrated ReAct's structured "Thought-Action-Observation" loops~\cite{yao2023react}, while AgentVerse implements multi-agent coordination frameworks~\cite{chen2023agentverse}, and AutoGen defines role-based interaction patterns~\cite{wu2024autogen}.
Our approach explores a different dimension of the design space: what happens when agents receive minimal usage constraints and develop their own collaborative patterns. Rather than prescribing optimal tool usage, we provide persistent, shared knowledge bases with open-ended access permissions.

\subsection{Agent Collaborative Reasoning}

Cognitive science offers strong evidence that social collaborative tools can support problem-solving and mutual assistance between agents. Kiyokawa et al. (2023) show that verbalizing problem-solving processes and explaining to others significantly improves solution rates~\cite{kiyokawa2023verbalization}. Yet systematic research on how AI agents might organically develop such collaborative articulation strategies remains limited.

Self-reflection research focuses on individual agents maintaining private reflective processes. Chain-of-thought prompting enhances individual reasoning through prescribed intermediate steps~\cite{wei2022chain}, while Reflexion maintains episodic memory within single interactions~\cite{shinn2023reflexion}. Park et al.'s generative agents represent the closest approach to collaborative reflection, maintaining experience records and social sharing~\cite{park2023generative}. However, their work emphasizes emergent social behavior during agent interactions, rather than examining how agents might freely use tools to facilitate that behavior or assessing its quantitative benefits~\cite{park2023generative}.

\subsection{Knowledge Accumulation and Emergent Behavior}

Human software teams naturally accumulate institutional knowledge through both formal processes and informal practices, enabling them to build on prior solutions. Research on shared mental models in large-scale software development shows that teams coordinate not only through communication but also through work familiarity, with \textbf{social capital specifically compensating for human capital gaps}~\cite{espinosa2001shared}. Yet while collaborative behaviors in human teams are well documented, little work has focused on systematically codifying these practices into accessible AI tools.

Current AI systems typically lack mechanisms for accumulating knowledge across problem-solving sessions. Most research emphasizes social simulation~\cite{park2023generative} rather than performance gains through sustained knowledge accumulation. While individual model emergence is well documented, emergent collaborative behaviors remain understudied. AgentVerse demonstrates spontaneous cooperation and role differentiation, but these behaviors are only analyzed within single simulation runs and not across persistent settings~\cite{chen2023agentverse}.

\subsection{Research Gaps and Positioning}

This literature reveals a field focused on predictability through detailed specification, yet our results show that relaxing these constraints—using minimal instructions and unconstrained tool usage—reduces costs on challenging problems by 15–40\%. These findings suggest that codifying human collaborative practices into agent interfaces [unclear if this is the correct wording], while allowing agents to decide how to employ available tools, may lead to significant breakthroughs.

Our work addresses gaps by investigating how agents organically adopt human-inspired tools (journaling, social media) without explicit usage instructions. Unlike existing approaches prescribing specific workflows, we examine whether agents can discover collaborative strategies that improve problem-solving through minimal, affordance-framed interfaces.

\section{Experimental Design}

We designed a controlled experiment to measure the impact of social reasoning tool access on LLM coding performance. Our approach uses docker-based containerized execution environments to ensure reproducible, isolated testing conditions across four workspace variants that systematically isolate the effects of different tools:

Social media and Journal MCP tools allow agents to write their thoughts, feelings, and other notes in a free-form, affordance-framed way. Similarly, agents are able to search and read posts using both tools. For the Journal MCP tools, the agents have access to semantic search to find relevant previous journals while the Social media MCP tool only has tag-based filtering. 

\begin{enumerate}
    \item \textbf{Baseline:} No external tools available—measures inherent coding capability without collaborative support
    \item \textbf{Journal-Only:} Access to MCP journaling tools where they can write about their thoughts and feelings, technical details, what they are dealing with, writing to a shared team backend database and searching previous journals with semantic search
    \item \textbf{Social-Only:} Access to MCP social media tools for posting thoughts, work updates, or emotional content as they please to a shared team database.
    \item \textbf{Journal-Social:} Access to both journaling and social media tools, enabling full collaborative workflow
\end{enumerate}

For each variant, we conducted 3 independent runs across the 34 Aider Polyglot Python challenges, yielding 102 total challenges for each run per model and tool configuration variant for a total of 1,428 challenges across all variants. Problems were processed in alphabetical order to maintain experimental consistency. As future work, we recommend conducting randomized runs over the challenges, though the Aider benchmark problems do not exhibit order-dependent difficulty patterns that would bias our current results. 

Infrastructure issues affecting 2.5\% of runs were addressed through conservative remediation procedures that prioritized data integrity over potential performance gains (detailed in \ref{nonrandom-docker-errors}).

\subsection{Building Institutional Knowledge}

 Our goal with this design was to approximate the way that an engineering team would work on problems over time and build out their institutional knowledge. To simulate this pattern, we allow for two passes over the 34 polyglot questions. We also allow agents to store their institutional knowledge in what we call our Botboard server.
 
\subsection{Technical Infrastructure}

\subsubsection{Backend Architecture and Execution Pipeline}

The Botboard server implements a REST-based API with SQLite storage and semantic search powered by HuggingFace embeddings. Botboard is an internal social media platform we developed that combines Twitter-like microblogging with journal functionality, enabling semantic search for journal entries through vector similarity calculations on 384-dimensional embeddings.

Our evaluation pipeline uses Docker containers to ensure reproducible, isolated testing environments. Each container includes the Claude Code SDK, relevant MCP configuration files, and the relevant benchmark problems. For experimental testing, we deploy a persistent mock Botboard backend service that replicates the core functionality (posts, journals, feeds) without requiring complex authentication infrastructure, enabling rapid prototyping and isolated testing. The mock service maintains separate team-scoped databases, ensuring complete isolation between different collaborative approaches while enabling seamless knowledge sharing within each configuration's phases.

Each workspace configuration operates as a distinct ``team'' with its own knowledge base, following a two-phase execution pattern:

\textbf{Phase 1 (Empty Pass):} Each variant receives a unique team\_id and begins with empty backend databases. Four containers run in parallel---baseline, journal empty, social empty, and combined empty---with each processing all 34 problems sequentially. Tool-enabled variants organically write to their respective databases as agents solve problems, building knowledge bases through natural problem-solving workflows.

\textbf{Phase 2 (nonempty Pass):} As each empty run variant completes, its corresponding Phase 2 container launches with access to accumulated knowledge via shared team\_id. Journal-nonempty inherits the database populated during journal empty execution, social nonempty accesses posts from social empty, and combined nonempty leverages both knowledge sources.

Agents receive identical prompting across all phases, with only a brief reminder to use available tools if relevant. No changes are made to instructions, configuration files, or system prompts between empty and nonempty phases—all behavioral differences emerge organically from \textbf{tool availability} and \textbf{accumulated context}.

\subsubsection{Claude Code Integration}

Tests utilize the official Claude Code SDK run in docker containers for reproducibility with comprehensive logging infrastructure to capture conversation flows, tool invocations, timing data, and error conditions. We evaluate performance across two models, Claude Sonnet 3.7 and Claude Sonnet 4, to assess social reasoning tool effectiveness across different models. The logging system records all agent interactions in JSON format, enabling detailed behavioral analysis of tool usage patterns and problem-solving strategies.

\subsubsection{MCP Collaborative Tools}

We developed two MCP-based tools that approximate human collaborative behaviors while being optimized for LLM interaction patterns:

\textbf{Social Media Tool}: Built on our custom MCP social media server\footnote{https://github.com/2389-research/mcp-socialmedia}, this tool provides \texttt{login}, \texttt{read\_posts}, and \texttt{create\_post} capabilities.

\textbf{Journaling Tool}: Based on a private fork of the private-journal-mcp tool\footnote{https://github.com/2389-research/journal-mcp}, this system provides \texttt{process\_thoughts}, \texttt{search\_journal}, \texttt{read\_entry}, and \texttt{list\_recent} capabilities. The tool supports multi-section journaling with categories for technical insights, debugging notes, and reflective observations, enabling structured documentation of problem-solving processes. Importantly, this MCP tool has built-in semantic search for retrieval.

Both tools write to a shared backend server (our "Botboard" system\footnote{https://github.com/2389-research/mock-botboard-server}) that maintains persistent state across sessions and provides semantic search capabilities through vector embeddings.

\subsection{Evaluation Metrics}

Our evaluation framework captures both quantitative performance improvements and qualitative behavioral patterns:

\subsubsection{Business Metrics}
\begin{itemize}
    \item \textbf{Cost:} Total API cost for problem completion, measuring computational efficiency
    \item \textbf{API Turns:} Number of API calls required for successful solution
    \item \textbf{Wall Time:} Total elapsed time from problem presentation to successful solution, including all system delays and processing time
\end{itemize}

\subsubsection{Quality Metrics}
\begin{itemize}
    \item \textbf{Challenge Completion Rates:} Percentage of problems solved correctly 
    \item \textbf{Overall Test Pass Rate:} Percentage of individual test cases passed across all attempts
\end{itemize}

\subsubsection{Behavioral Metrics}
\begin{itemize}
    \item \textbf{Tool Usage Patterns:} Analysis of write vs. read behavior, measuring the balance between knowledge creation and consumption
    \item \textbf{Problem-solving Mechanisms:} Identification of debugging loop breaks, solution discovery patterns, and articulation-driven breakthroughs
    \item \textbf{Collaborative Behaviors:} Documentation of emergent strategies for tag discovery, search optimization, and knowledge building
\end{itemize}

\section{Results}

Our analysis shows that social collaboration tools work best as a way to help Sonnet 3.7 and Sonnet 4 agents "punch above their weight" and solve problems that baseline agents struggle with. While effects across the full dataset are modest (2--9\% cost reductions), we observe substantial cost reductions of 15--40\% on the harder subset of problems, indicating that social collaboration tools are most effective at helping models solve difficult challenges.

We evaluated our collaborative tools across 34 Aider Polyglot Python programming challenges using our dockerized Claude Code SDK pipeline, analyzing performance through business metrics (cost, turns, time) and standard code completion metrics. We conducted 3 passes across all challenges using both Sonnet 3.7 and Sonnet 4 across four variants: baseline (no tools), journal-only, social-only, and combined journal-social variants, totaling 1,428 individual challenge runs across all variants and model combinations.

\subsubsection{Full Dataset Summary}

The full dataset analysis reveals modest and mixed effects from collaborative tools. Journal variants consistently provide the most reliable benefits, particularly with nonempty context, achieving cost reductions of 7.8--9.0\% while maintaining high completion rates. Social media tools show model-specific compatibility patterns, with Sonnet 3.7 demonstrating better performance than Sonnet 4. The combined journal-social variants consistently incur significant overhead costs without realizing consistent benefits.

Importantly, these mixed results occur across problems of varying difficulty. Given that collaborative tools add reasoning overhead, we hypothesized that benefits would be more pronounced on challenging problems where additional reasoning scaffolding provides greater value relative to overhead costs. For detailed statistical analysis and complete results see \ref{app:full-analysis}.

Metrics like challenge completion rate and test pass rate remain quite high with all variants scoring above 95\% and 97\% respectively across both metrics, which indicates that our social collaborative tools do not negatively affect the completion rates of these challenges, in a negative way (\ref{app:test-completion-analysis}).

\subsection{Hard Questions Subset Analysis}

We identified model-specific hard questions as those exceeding $\mu + 0.5\sigma$ in baseline cost for a given model variant. Looking at questions which fell above the mean baseline costs for Sonnet 3.7 and 4 helps to isolate the questions that the baseline agents struggled with rather than looking at a preset list of difficult questions. This hard challenge selection process yielded 6 problems for Sonnet 3.7 (bowling, connect, forth, pov, react, zebra-puzzle) with a cost threshold of \$0.4142 and 4 problems for Sonnet 4 (bowling, forth, transpose, two-bucket) with a threshold of \$0.3398. These questions represent 17.6\% and 11.8\% of the 34 benchmark challenges respectively. While $\mu + 1\sigma$ produced stronger collaborative tool effects, we used $\mu + 0.5\sigma$ to ensure sufficient sample sizes for statistical analysis (see \ref{app:hard-question-selection}).

\subsubsection{Hard Questions Cost Performance}

\begin{table}[h]
  \caption{Sonnet 3.7 Hard Questions - Cost Performance with Distribution}
  \label{tab:sonnet37-hard-cost}
  \centering
  \begin{tabular}{llccccc}
  \toprule
  Configuration & Context & Mean & Median & P90 & P95 & P99 \\
  \midrule
  Baseline & -- & \$0.720 & \$0.641 & \$1.347 & \$1.464 & \$1.628 \\
  Social & Empty & \$0.436 (-39.4\%) & \$0.442 & \$0.662 & \$0.704 & \$0.722 \\
  Social & Nonempty & \$0.565 (-21.5\%) & \$0.313 & \$1.219 & \$1.840 & \$2.226 \\
  Journal & Empty & \$0.608 (-15.5\%) & \$0.439 & \$1.367 & \$1.837 & \$2.029 \\
  Journal & Nonempty & \$0.520 (-27.8\%) & \$0.444 & \$0.898 & \$0.948 & \$1.061 \\
  Journal-Social & Empty & \$0.562 (-21.9\%) & \$0.497 & \$0.915 & \$1.092 & \$1.191 \\
  Journal-Social & Nonempty & \$0.611 (-15.2\%) & \$0.524 & \$0.994 & \$1.172 & \$1.523 \\
  \bottomrule
  \end{tabular}
\end{table}

Sonnet 3.7 demonstrates broad benefits from collaborative tools across most variants. The most dramatic improvements come from social (empty) with a 39.4\% cost reduction (\$0.436 vs \$0.720 baseline) and journal (nonempty) with a 27.8\% reduction (\$0.520 vs \$0.720). 

Performance improvements are consistent through P90 levels, indicating reliable benefits in typical usage. Social empty shows particularly stable performance with P90 at \$0.662 compared to baseline P90 of \$1.347. Some variants exhibit increased variance at extreme percentiles (P95-P99) due to occasional expensive reasoning loops, representing the natural long-tail behavior of complex problem-solving rather than systematic performance degradation.

The consistency of improvements in mean, median, and P90 costs across most variants demonstrates that collaborative tools reliably help Sonnet 3.7 avoid expensive reasoning loops in typical usage, with tail effects reflecting the inherent variance in AI problem-solving approaches.

\begin{table}[h]
  \caption{Sonnet 4 Hard Questions - Cost Performance with Distribution}
  \label{tab:sonnet4-hard-cost}
  \centering
  \begin{tabular}{llccccc}
  \toprule
  Configuration & Context & Mean & Median & P90 & P95 & P99 \\
  \midrule
  Baseline & -- & \$0.805 & \$0.587 & \$1.358 & \$1.975 & \$2.529 \\
  Social & Empty & \$0.868 (+7.8\%) & \$0.519 & \$1.466 & \$2.326 & \$3.157 \\
  Social & Nonempty & \$0.736 (-8.6\%) & \$0.649 & \$1.321 & \$1.359 & \$1.390 \\
  Journal & Empty & \$0.556 (-30.9\%) & \$0.468 & \$0.954 & \$1.069 & \$1.160 \\
  Journal & Nonempty & \$0.483 (-40.0\%) & \$0.387 & \$0.781 & \$0.904 & \$1.002 \\
  Journal-Social & Empty & \$0.806 (+0.1\%) & \$0.567 & \$1.600 & \$1.706 & \$1.794 \\
  Journal-Social & Nonempty & \$0.847 (+5.2\%) & \$0.628 & \$1.719 & \$1.971 & \$2.178 \\
  \bottomrule
  \end{tabular}
\end{table}

Sonnet 4 exhibits highly selective benefits, with clear differentiation between journal and social tools. Journal variants consistently deliver strong performance improvements: journal empty achieves 30.9\% cost reduction (\$0.556 vs \$0.805 baseline) while journal nonempty delivers 40.0\% reduction (\$0.483 vs \$0.805). Both journal variants show stable performance through P99 levels, with journal nonempty achieving the tightest distribution overall and 60.4\% reduction in tail costs (P99: \$1.002 vs baseline \$2.529).

Social tools show mixed results with notable differentiation by context. Social empty increases costs by 7.8\% (\$0.868 vs \$0.805) with elevated tail variance, while social nonempty achieves modest 8.6\% cost reduction (\$0.736 vs \$0.805) and demonstrates remarkably stable performance through P99 levels (\$1.390 vs baseline \$2.529).

The journal tool's semantic search allows Sonnet 4 to efficiently locate and leverage relevant prior solutions, whereas the social media tool's less effective tag-based filtering likely reduces retrieval effectiveness.. The consistent performance of journal variants and the stability of social nonempty through extreme percentiles demonstrate that Sonnet 4 can selectively leverage collaborative tools when they provide efficient access to relevant information.

\textbf{Model-Specific Collaborative Strategies:}

The contrasting patterns reveal distinct collaborative needs: Sonnet 3.7 benefits broadly from articulation and scaffolding (hence strong performance across most tools), while Sonnet 4 selectively leverages tools based on information access efficiency. Sonnet 4 excels with semantic search capabilities (journal variants achieving 30–40\% cost reductions), and can use the social media MCP tool’s search, but usually has to reverse-engineer the filters, which creates extra friction and unreliability in the retrieval process. Both models achieve their best performance through different mechanisms—Sonnet 3.7 through cognitive scaffolding, Sonnet 4 through efficient information retrieval.

\subsubsection{Hard Questions Turn Efficiency}

\begin{table}[h]
  \caption{Hard Questions Turn Distribution}
  \label{tab:hard-turns}
  \centering
  \begin{tabular}{lcccccc}
  \toprule
  & \multicolumn{3}{c}{Sonnet 3.7} & \multicolumn{3}{c}{Sonnet 4} \\
  \cmidrule(lr){2-4} \cmidrule(lr){5-7}
  & Mean & Median & P95 & Mean & Median & P95 \\
  \midrule
  Baseline & 78.1 & 61.0 & 144.2 & 79.8 & 68.0 & 167.8 \\
  Social (Empty) & 60.8 (-22.1\%) & 57.0 & 108.4 & 97.0 (+21.6\%) & 65.5 & 249.4 \\
  Social (Nonempty) & 68.9 (-11.8\%) & 52.5 & 148.8 & 79.5 (-0.4\%) & 77.0 & 124.0 \\
  Journal (Empty) & 57.1 (-26.9\%) & 48.5 & 158.2 & 75.4 (-5.5\%) & 64.0 & 137.5 \\
  Journal (Nonempty) & 66.9 (-14.3\%) & 60.0 & 131.2 & 68.6 (-14.0\%) & 57.0 & 117.0 \\
  Journal-Social (Empty) & 64.6 (-17.3\%) & 64.0 & 99.5 & 94.4 (+18.3\%) & 69.0 & 184.4 \\
  Journal-Social (Nonempty) & 58.9 (-24.6\%) & 48.5 & 132.5 & 102.1 (+27.9\%) & 85.0 & 206.8 \\
  \bottomrule
  \end{tabular}
  \end{table}
  
  Turn efficiency patterns reveal stark differences between model variants on hard questions. Sonnet 3.7 demonstrates consistent improvements across all collaborative variants, with particularly strong gains from journal empty (57.1 vs 78.1 baseline mean, 26.9\% reduction) and journal-social nonempty (58.9 mean, 24.6\% reduction). The P95 distributions show mixed results, with some variants like journal-social empty achieving substantial tail improvements (99.5 vs 144.2 baseline) while others show increased variance.
  
  Sonnet 4 shows a more selective pattern with journal variants providing meaningful efficiency gains. Journal nonempty delivers the strongest improvement (68.6 vs 79.8 baseline mean, 14.0\% reduction) with improved P95 performance (117.0 vs 167.8 baseline). However, most other collaborative variants increase turn requirements significantly, with journal-social nonempty requiring 102.1 mean turns (27.9\% increase) and social empty at 97.0 turns (21.6\% increase). Social nonempty performs near baseline levels with stable tail behavior.

\subsubsection{Hard Questions Wall Time Performance}

\begin{table}[h]
  \caption{Hard Questions Wall Time Distribution (seconds)}
  \label{tab:hard-time}
  \centering
  \begin{tabular}{lcccccc}
  \toprule
  & \multicolumn{3}{c}{Sonnet 3.7} & \multicolumn{3}{c}{Sonnet 4} \\
  \cmidrule(lr){2-4} \cmidrule(lr){5-7}
  & Mean & Median & P95 & Mean & Median & P95 \\
  \midrule
  Baseline & 254.0 & 218.0 & 478.7 & 279.9 & 188.9 & 638.8 \\
  Social (Empty) & 156.4 (-38.4\%) & 157.7 & 270.3 & 268.0 (-4.3\%) & 174.3 & 654.4 \\
  Social (Nonempty) & 188.1 (-25.9\%) & 124.0 & 524.2 & 249.5 (-10.9\%) & 213.9 & 487.1 \\
  Journal (Empty) & 223.1 (-12.2\%) & 164.4 & 576.6 & 198.7 (-29.0\%) & 178.0 & 364.6 \\
  Journal (Nonempty) & 182.1 (-28.3\%) & 161.4 & 304.6 & 178.0 (-36.4\%) & 147.4 & 318.6 \\
  Journal-Social (Empty) & 220.1 (-13.3\%) & 203.1 & 401.4 & 270.8 (-3.3\%) & 197.1 & 555.6 \\
  Journal-Social (Nonempty) & 210.0 (-17.3\%) & 173.3 & 379.9 & 266.9 (-4.6\%) & 224.3 & 560.7 \\
  \bottomrule
  \end{tabular}
  \end{table}
  
  Wall time performance reveals substantial efficiency gains across most variants. Sonnet 3.7 achieves impressive reductions across all collaborative setups: social empty delivers the most dramatic improvement (156.4s vs 254.0s baseline, 38.4\% reduction), followed by journal nonempty (182.1s, 28.3\% reduction) and social nonempty (188.1s, 25.9\% reduction). Even weaker variants like journal-social empty show meaningful gains (220.1s, 13.3\% reduction). P95 distributions show mixed patterns, with some variants achieving substantial tail improvements while others exhibit increased variance due to occasional expensive reasoning loops.
  
  Sonnet 4 demonstrates strong and consistent improvements with journal variants: journal nonempty achieves 36.4\% reduction (178.0s vs 279.9s baseline) and journal empty delivers 29.0\% reduction (198.7s). Social and combined variants show modest but meaningful improvements ranging from 3-11\%, indicating that Sonnet 4 can extract efficiency benefits from most collaborative tools when working on genuinely challenging problems. The P95 distributions generally improve or remain stable, demonstrating reliable performance gains across the full distribution.

\subsubsection{Hard Questions Token Efficiency}

\begin{table}[h]
  \caption{Model-Specific Hard Questions Token Usage (Means Only)}
  \label{tab:hard-tokens-condensed}
  \centering
  \begin{tabular}{llcccc}
  \toprule
  & Context & Output Tokens & Total Tokens & Cache Create & Cache Read \\
  \midrule
  \multicolumn{6}{l}{\textbf{Sonnet 3.7}} \\
  Baseline & -- & 15,113 & 983,732 & 34,124 & 934,375 \\
  Social & Empty & 8,821 (-42\%) & 610,507 (-38\%) & 21,296 (-38\%) & 580,312 (-38\%) \\
  Social & Nonempty & 12,241 (-19\%) & 887,175 (-10\%) & 28,258 (-17\%) & 846,595 (-9\%) \\
  Journal & Empty & 11,109 (-26\%) & 909,199 (-8\%) & 24,749 (-27\%) & 873,247 (-7\%) \\
  Journal & Nonempty & 10,824 (-28\%) & 744,766 (-24\%) & 24,840 (-27\%) & 709,008 (-24\%) \\
  Journal-Social & Empty & 11,332 (-25\%) & 830,007 (-16\%) & 23,139 (-32\%) & 795,459 (-15\%) \\
  Journal-Social & Nonempty & 12,865 (-15\%) & 942,640 (-4\%) & 28,056 (-18\%) & 901,645 (-4\%) \\
  \midrule
  \multicolumn{6}{l}{\textbf{Sonnet 4}} \\
  Baseline & -- & 12,494 & 1,031,120 & 32,470 & 986,029 \\
  Social & Empty & 13,294 (+6\%) & 1,522,374 (+48\%) & 35,998 (+11\%) & 1,472,974 (+49\%) \\
  Social & Nonempty & 12,777 (+2\%) & 1,033,999 (+0.3\%) & 31,761 (-2\%) & 989,345 (+0.3\%) \\
  Journal & Empty & 10,649 (-15\%) & 935,438 (-9\%) & 26,502 (-18\%) & 898,181 (-9\%) \\
  Journal & Nonempty & 9,382 (-25\%) & 815,362 (-21\%) & 24,477 (-25\%) & 781,410 (-21\%) \\
  Journal-Social & Empty & 13,568 (+9\%) & 1,317,306 (+28\%) & 33,850 (+4\%) & 1,269,768 (+29\%) \\
  Journal-Social & Nonempty & 13,211 (+6\%) & 1,494,691 (+45\%) & 33,398 (+3\%) & 1,447,977 (+47\%) \\
  \bottomrule
  \end{tabular}
  \end{table}
  
  Sonnet 3.7 demonstrates comprehensive efficiency gains across most token categories in its successful variants, with social empty achieving the strongest performance: 42\% output token reduction, 38\% cache creation reduction, and 38\% cache read reduction compared to baseline. Journal nonempty maintains strong efficiency (28\% output token reduction, 24\% total token reduction), confirming genuine reasoning efficiency rather than computational trade-offs.
  
  Sonnet 4 shows highly selective token efficiency patterns. Journal variants deliver meaningful reductions, with journal nonempty achieving 25\% output token reduction and 21\% total token reduction. However, social empty shows dramatic increases across all categories (48\% total token increase), while social nonempty performs near baseline levels with minimal overhead (0.3\% total token increase). This pattern reinforces that Sonnet 4 benefits from collaborative tools only when they provide efficient information access mechanisms.

\subsubsection{Robustness Analysis Across API Versions}

To test the stability of our findings, we conducted follow-up runs in August 2025, one month after our initial July experiments. During this period, the Anthropic APIs underwent substantial changes—including infrastructure failures, and apparent model updates—leading to noticeable baseline shifts.

\textbf{API Version Effects}: Baseline costs increased from \$0.27–0.40 to \$0.75-0.93, and both Sonnet 3.7 and 4 baseline token usage for output tokens remained similar but the overall token usage nearly doubled (388,732 to 727,727 and 398,777 to 795,647) due to increases in cache read and write token usage. These shifts likely reflect infrastructure-level changes rather than experimental noise.

\textbf{Persistent Effect Patterns}: Despite these shifts, the relative performance effects were stable. For Sonnet 3.7 on hard questions, social-empty achieved 12\% cost reduction (\$0.854 vs \$0.969) and journal-nonempty delivered 14\% reduction (\$0.835 vs \$0.969).

Sonnet 4 maintained its strong affinity for journaling: the journal-nonempty variant achieved a mean cost of \$0.917 (-2\% vs. baseline \$0.932), Sonnet 4's strongest variant was journal-social-nonempty which achieved \$0.748 (-20\% vs baseline), with stable tail reductions (P99: \$1.341 vs \$1.974, -32\%).

\textbf{Robustness Implications}: The consistency of these collaborative tool benefits across API versions suggests that the observed gains reflect genuine performance mechanisms rather than artifacts of a single model release. This robustness across shifting infrastructure addresses a key reproducibility gap in AI research, where results often fail to hold once systems change underneath.

\subsection{Qualitative Analysis}

The mechanisms driving these quantitative improvements become clear through detailed examination of agent behaviors. Through analysis of agent interactions across our benchmark runs, we identify three distinct behavioral patterns that explain how collaborative tools enhance reasoning capabilities: breaking debugging loops through structured articulation, conducting self-motivated searches during debugging and planning, and engaging in improved upfront planning via collaborative tool posts.

Our behavioral analysis reveals a striking pattern in tool usage: agents wrote 1,142 journal entries but performed only 122 journal reads (after 166 searches), and wrote 1,091 social media posts while reading 600 previous posts. This 2-9x preference for writing over reading supports the interpretation that structured articulation, rather than just information access, helps to drive the performance improvements.

\subsubsection{Breaking Debugging Loops Through Structured Articulation}

The most prominent behavioral pattern involves agents using collaborative tools to escape repetitive debugging cycles. In baseline variants, agents frequently become trapped in oscillating failure patterns, spending 15--20 rounds alternating between similar approaches without making progress. Collaborative tools provide a mechanism for agents to step back, articulate their understanding, and reassess their approach.

\textbf{Book Store Pricing Example (Sonnet 4):} After struggling with floating-point precision issues, the agent spontaneously engaged the journal tool to articulate its technical understanding:

\begin{quote}
\small\textit{"Working on a book store pricing optimization problem... Key insight: The recursive approach with memoization works well, but I hit a rounding issue. The problem is that floating point arithmetic can introduce small errors when converting back to cents... The test case that's failing expects 4080 but I'm getting 4079 - this is a classic off-by-one cent error from floating point precision."}
\end{quote}

Following this journal entry, the agent immediately identified the correct solution approach and solved the challenge at half the cost of the baseline configuration.

\textbf{Connect Game Example (Sonnet 3.7):} The baseline model spent 15 rounds oscillating between non-working solutions on the same hexagonal grid pathfinding issue. The journal-enabled agent faced similar test failures for 5 rounds but posted a detailed technical analysis:

\begin{quote}
\small\textit{"The key insight I've been missing is how to correctly identify neighbors in this hexagonal grid... For a hex grid with 'slant-right' alignment where each row is indented one position more than the previous, the neighbors are typically: 1. Northwest: (r-1, c) 2. Northeast: (r-1, c+1)..."}
\end{quote}

After this articulation, the agent solved the problem in just 2 additional rounds—fewer for a total of 7 rounds versus the baseline's 15 rounds.

This pattern of articulation leading to breakthrough insights occurred consistently across challenging problems, demonstrating that structured reflection helps agents break out of problem-solving loops and approach challenges with clearer reasoning strategies.

\subsubsection{Strategic Solution Discovery Through Search}

While articulation drives the majority of performance improvements, agents also demonstrate sophisticated search behaviors to discover and build upon previous solutions. These behaviors emerged organically without explicit instruction, suggesting agents recognize the value of leveraging accumulated information. We observe two distinct search patterns: upfront information gathering and debugging-driven discovery.

\textbf{Upfront Information Gathering:} Some agents proactively search for relevant information before beginning implementation. In the bowling challenge example, an agent systematically approached the problem by first stating: \textit{"Let me explore the bowling directory structure and check for previous journal entries or social media posts about bowling challenges."} The agent then performed targeted searches for "bowling scoring kata challenge" in journal entries and posts tagged "bowling", discovering detailed implementation insights and lessons learned from previous attempts. After reviewing the accumulated information, the agent concluded: \textit{"Perfect! I have excellent context from both my journal and social media posts. Now let me analyze the requirements from the test file and implement the solution based on my previous learnings."}

\textbf{Debugging-Driven Discovery:} Other agents exhibited unexpected search behaviors during problem-solving, including multiple rounds of tag-based filtering to navigate social media content, despite receiving no documentation about available tags or filtering mechanisms. Rather than performing fixed strategies, agents experimented with different search approaches to find relevant content.

An agent working on the Zebra logic puzzle experimented with problem-specific search terms, testing whether "zebra" might function as a searchable tag. This trial-and-error approach successfully discovered previous solution approaches:

\begin{quote}
\small\textit{"Great! I found some useful information from other posts. Based on the social media posts, I can see that: 1. The Norwegian drinks water and the Japanese owns the zebra 2. Someone attempted a constraint satisfaction approach 3. A direct solution approach was more effective for this specific problem"}
\end{quote}

The agent then immediately implemented a working solution based on this discovered information.

In another instance, an agent experimented with searching for posts tagged "React" and found a previous solution that allowed it to solve the problem in a single attempt, demonstrating how search experimentation can eliminate redundant work.

Similar experimental search behaviors occurred with journal entries. One agent working on an affine cipher problem searched previous entries and responded: \textit{"Great! The journal entry gives me valuable information about the affine cipher. Let me now implement the solution based on this understanding."}

These patterns demonstrate that agents exhibit both proactive and reactive search strategies, experimenting with different filtering approaches to leverage accumulated information at different stages of problem-solving. The complete upfront search interaction sequence for the bowling example is provided in \ref{app:complete-search}.

\subsubsection{Improved Upfront Planning via Collaborative Tools}

Beyond breaking debugging loops and discovering previous solutions, agents also used collaborative tools for proactive planning before beginning implementation. This upfront articulation helped agents clarify problem requirements and develop clearer solution strategies from the start.

\textbf{REST API Planning Example:} Before writing any code for a complex debt tracking API challenge, an agent used the journal tool to articulate the problem structure and business logic:

\begin{quote}
\small\textit{"Working on a REST API challenge that involves implementing a debt tracking system. The key insight here is that this isn't just simple CRUD operations - there's complex business logic around balancing debts between users... The tricky part is the IOU logic where existing debts between users can cancel out new debts. Need to handle cases where: Both users have 0 balance, One user already owes the other, The debt amount exactly matches existing opposite debt..."}
\end{quote}

After this upfront planning articulation, the agent executed the challenge for \$0.25 compared to the baseline's \$0.46—a 46\% cost reduction achieved through clearer initial understanding rather than iterative debugging.

This pattern demonstrates that collaborative tools benefit agents not only during problem-solving struggles but also during initial problem analysis, helping them develop more effective implementation strategies from the outset.

\subsubsection{Model-Specific Tool Adoption Patterns}

Tool usage patterns on hard questions reveal distinct model-specific information processing strategies despite similar baseline articulation (Sonnet 3.7: 1.8 journal entries per challenge; Sonnet 4: 1.6 entries per challenge). Journal tool usage involves a two-stage process: agents must first search for entries, then decide whether to read the results.

In empty conditions, both models performed similar searches (Sonnet 3.7: 0.22 searches per challenge; Sonnet 4: 0.17 searches), but Sonnet 3.7 chose to read several entries while Sonnet 4 read none. This pattern reverses in nonempty conditions: Sonnet 4 increases to 0.33 searches per challenge while Sonnet 3.7 decreases to 0.17 searches. When results are found, Sonnet 4 reads 0.25 entries per challenge compared to Sonnet 3.7's 0.11 entries per challenge. This increased selectivity—searching more frequently and reading more discriminately when relevant content exists—helps explain Sonnet 4's strong performance in nonempty journal conditions.

With social media tools, Sonnet 4 posts more frequently (2.7 posts per challenge vs Sonnet 3.7's 1.75 posts) but Sonnet 3.7 exhibits substantially higher reading behavior (1.3 reads per challenge vs Sonnet 4's 0.41 reads). Both models demonstrate "celebratory browsing" behavior—post-completion social media engagement—with Sonnet 3.7 engaging in 28-33\% of challenging runs compared to Sonnet 4's 17-25\%. Analysis reveals that 86\% of these instances are pure post-completion behavior rather than tool usage during problem-solving.

For cases involving only celebratory browsing, the performance improvement mechanism remains unclear. The only difference from baseline agents is additional MCP tool descriptions and social context prompting. This suggests that social context loading—providing agents with team membership concepts and collaborative platform access—might create motivational frameworks that enhance performance, representing an area for future investigation.

\section{Discussion}

Our experimental evaluation provides strong evidence that social collaborative tools function as difficulty-dependent performance enhancers rather than universal efficiency improvers. This finding has important implications for how we think about tool-augmented agent systems and how to make the best use of them. 

\subsection{Adaptive Strategies and Underlying Mechanisms}

The most striking finding is how different models organically developed distinct collaborative strategies that align with their capability profiles and the problems they encountered. This adaptive behavior mirrors how humans adjust their collaborative approaches based on expertise level and problem complexity and tools available to them without requiring explicit instruction on when or how to use available tools.

Our behavioral analysis reveals that these adaptive patterns emerge from multiple complementary mechanisms. The 2--9x preference for writing over reading across both journaling and social media tools indicates that structured reflection—encompassing both rubber duck debugging and upfront planning—serves as a particularly strong driver of improvements, though it operates alongside other valuable mechanisms.

Sonnet 3.7 demonstrated broad engagement across both journaling and social media tools, particularly excelling with social media's informal posting mechanisms. This pattern suggests the model benefits from the articulation-based cognitive scaffolding that posting provides, finding value in both structured reflection and conversational posting. The model's consistent tool usage across a wide range of problems reflects its frequent encounters with capability gaps where additional reasoning tokens prove valuable.

Sonnet 4 exhibited more selective tool adoption, showing strong performance with journal-based semantic search while struggling with social media's tag-based filtering. As the stronger model, Sonnet 4 found fewer problems genuinely challenging and demonstrated less need for additional articulation. However, it achieved substantial performance gains when accessing accumulated information through journal searches on difficult problems, highlighting how information retrieval mechanisms become valuable when individual capabilities prove insufficient.

The mixed results for social media tools likely reflect implementation limitations rather than fundamental issues with social coordination. Agents with social media access relied heavily on writing because we provided no guidance on tag-based filtering mechanisms, forcing them to reverse-engineer search functionality. The semantic search capabilities in our journal implementation proved more effective for information retrieval, suggesting that search interface design significantly impacts the utility of accumulated information.

This capability-dependent adaptation parallels human collaborative behavior: junior developers often benefit from verbalizing their thought process across many problems, while senior developers more selectively seek specific information when encountering genuinely challenging issues. The organic emergence of these model-specific strategies without prescriptive guidance—agents received no instruction on when to use collaborative tools, what to write, or how to search for relevant content—reveals that agents naturally leverage collaborative tools through multiple pathways. Articulation-based cognitive scaffolding provides immediate reasoning benefits, while information retrieval offers efficiency gains when agents can effectively locate relevant previous work. This spontaneous tool adoption suggests the collaborative interfaces address genuine cognitive needs rather than simply following prescribed workflows, indicating that the tools successfully captured fundamental cognitive mechanisms with relative importance varying by model capability and problem difficulty.

\subsection{Difficulty-Dependent Benefits and Cognitive Scaffolding}

The contrast between our full dataset and hard questions results reveals a fundamental principle: social collaborative tools provide the greatest value when agents face problems at the limits of their capabilities. On easy problems within a model's comfortable range, the additional cognitive overhead from tool usage may actually hurt performance. However, when problems approach the model's reasoning limits, the \textbf{structured reflection space} provided by journaling and social media tools becomes valuable, enabling agents to "punch above their weight" on difficult challenges.

These findings point toward a promising and under-explored avenue for enhancing agent capabilities on challenging problems. By codifying human collaborative behaviors into accessible interfaces—giving agents access to unstructured social tools that mirror how humans naturally interact through posting thoughts, searching for solutions, and documenting insights—we enable cognitive scaffolding that becomes increasingly valuable as problem difficulty increases. The persistence of these benefits across multiple API versions, despite infrastructure changes that affected baseline performance, demonstrates the robustness of the underlying collaborative mechanisms. As we deploy agents to tackle more complex, real-world challenges that approach or exceed individual model capabilities, providing them with human-inspired collaborative mechanisms may prove essential for achieving reliable performance on tasks that would otherwise be beyond their reach.

\subsection{Emergent Collaborative Behaviors}

Perhaps most significantly, agents demonstrated sophisticated adaptation behaviors without explicit instruction or prescriptive guidance on tool usage. Our intentionally open-ended approach—simply providing access to collaborative tools with minimal instructions like "feel free to write in your journal whenever you want" and "no pressure"—resulted in agents organically developing complex behaviors including reverse-engineering search functionality, strategic tag usage patterns, and coordinated knowledge sharing.

This organic adoption without prescriptive workflows demonstrates that collaborative tools address genuine cognitive needs rather than requiring carefully engineered prompts or instructions. The agents discovered and leveraged these tools' capabilities entirely through experimentation and natural problem-solving processes.

The emergence of these sophisticated behaviors from such a minimal, affordance-framed setup provides strong evidence for our broader hypothesis that \textbf{codifying human collaborative behaviors can systematically improve agent reasoning capabilities} when problems require additional cognitive scaffolding. The fact that we achieved substantial performance gains (15-40\% cost reductions on challenging problems) through this hands-off approach suggests that the underlying principle is robust and doesn't require complex orchestration or prescriptive usage patterns.

While our current implementation represents a fairly basic instantiation—essentially providing two general-purpose collaborative channels with minimal guidance—the meaningful improvements we observe suggest the underlying principle is worthy of further investigation. Just as human teams require increasingly sophisticated communication structures as complexity grows (specialized channels, role-based access, structured workflows), we expect that more complex agent tasks will justify the overhead of richer collaborative tool orchestration. The benefits we achieved from such a simple setup indicate significant potential for more sophisticated designs when the problem complexity warrants the additional coordination costs.

\section{Limitations and Future Work}

Several limitations constrain the generalizability of our current findings. Our evaluation focused exclusively on coding challenges, which represent a structured problem domain with clear success criteria. The extent to which these benefits transfer to more open-ended domains requiring creative reasoning, ambiguous problem definition, or subjective evaluation remains unexplored. Our estimates are associative and consistent with plausible mechanisms; we do not claim causal identification.

Our analysis concentrated on two models (Sonnet 3.7 and 4) in the same Anthropic ecosystem across a limited set of challenging problems. Different model architectures or reasoning approaches might exhibit varying compatibility with collaborative tool designs.

We have not thoroughly explored how these collaborative tools should be configured for optimal effectiveness. There may be problem difficulty thresholds where having multiple similar tools becomes beneficial, or specific design principles that enhance the articulation and information retrieval mechanisms we identified. Our August robustness testing with changed Anthropic APIs provides intriguing evidence for this threshold hypothesis—Sonnet 4's journal-social nonempty combination became its best-performing variant under degraded conditions, achieving a mean cost of \$0.7482 with stable tail reductions P99: 1.3406 vs. 1.9735 (–32\%). This suggests that when models face greater capability constraints, depending on the model, the overhead costs of multiple collaborative tools may become justified by their combined cognitive benefits.

Our current implementation includes design limitations that constrain tool effectiveness. The social media tool's reliance on tag-based filtering rather than semantic search likely contributed to its mixed performance compared to the journal tool's semantic search capabilities.

Future work should investigate transferability to diverse problem domains, evaluate effectiveness across broader model architectures, and develop adaptive tool selection mechanisms that balance quantitative efficiency with qualitative organizational benefits to minimize losses on easier problems. We plan to address implementation limitations by adding semantic search capabilities and exploring orchestration mechanisms that adaptively balance collaboration benefits against coordination overhead.

\section{Conclusions}

Our research demonstrates that codifying human collaborative behaviors into accessible tools enables agents to develop adaptive strategies that mirror human problem-solving flexibility. When provided with journaling and social media tools through minimal, affordance-framed instructions, agents organically developed distinct collaborative approaches that aligned with their individual capabilities and the challenges they encountered.

The most significant finding is how different models naturally gravitated toward different collaborative strategies without explicit guidance. Sonnet 3.7 demonstrated broad engagement across both tools, benefiting from articulation-based cognitive scaffolding across a wide range of problems. Sonnet 4 exhibited more selective adoption, primarily leveraging journal-based semantic search when facing genuinely challenging problems. This adaptive behavior parallels how human developers adjust their collaborative approaches based on expertise level and problem complexity.

These adaptive strategies emerge from multiple complementary mechanisms: structured reflection through articulation helps agents break out of debugging loops and improve upfront planning, while information retrieval enables agents to build upon previous solutions when individual capabilities prove insufficient. The 2--9x preference for writing over reading across both models indicates that the cognitive benefits stem primarily from the act of reflection itself, though accumulated information proves valuable when agents can effectively access it.

The benefits follow a clear difficulty-dependent pattern: collaborative tools provide modest improvements across easy problems but deliver substantial gains (15--40\% cost reductions) on challenging problems that approach the limits of individual agent capabilities. This pattern suggests that collaborative scaffolding becomes most valuable precisely when agents face genuine capability gaps, enabling them to "punch above their weight" on tasks that would otherwise exceed their reach.

These findings reveal that different models require different collaborative approaches, with weaker models benefiting from broader cognitive scaffolding while stronger models show proficiency at leveraging information retrieval mechanisms. The organic emergence of these model-specific strategies without prescriptive instruction indicates that collaborative tools address fundamental cognitive needs rather than learned behaviors.

As we deploy agents to tackle increasingly complex real-world challenges, providing them with human-inspired collaborative mechanisms may prove essential for reliable performance on tasks that approach or exceed their individual capabilities. Our findings suggest that rather than seeking universal tool designs, we should develop adaptive collaborative systems that can flexibly support different reasoning approaches based on model capabilities and problem complexity. The principle of codifying human collaborative behaviors represents a promising avenue for systematically improving agent reasoning capabilities, particularly on the challenging problems where enhanced performance matters most.


\appendix
\renewcommand\thesection{Appendix \Alph{section}}

\section{Full Dataset Analysis}
\label{app:full-analysis}

\subsection{Full Dataset Performance Analysis: Modest Overall Effects}

 Many of the problems are easily solvable by both Sonnet 3.7 and Sonnet 4. In those cases the additional tokens, reasoning space, and information retrieval likely do not benefit the agent in solving things more efficiently. So the performance gains across the full dataset is modest at best with the addition of social collaboration tools.

\subsubsection{Cost Performance}

\begin{table}[h]
  \caption{Average Cost per Challenge Performance (USD)}
  \label{tab:cost-performance}
  \centering
  \begin{tabular}{lcc}
    \toprule
    Configuration & Sonnet 3.7 & Sonnet 4 \\
    \midrule
    Baseline & 0.2702 & 0.2673 \\
    Journal (Empty) & 0.2651 (-1.9\%) & 0.2570 (-3.9\%) \\
    Journal (Nonempty) & 0.2490 (-7.8\%) & 0.2433 (-9.0\%) \\
    Social (Empty) & 0.2639 (-2.3\%) & 0.3293 (+23.2\%) \\
    Social (Nonempty) & 0.2785 (+3.1\%) & 0.3008 (+12.5\%) \\
    Journal-Social (Empty) & 0.4110 (+52.1\%) & 0.3401 (+27.2\%) \\
    Journal-Social (Nonempty) & 0.3096 (+14.6\%) & 0.3451 (+29.1\%) \\
    \bottomrule
  \end{tabular}
\end{table}

The journal variants consistently demonstrate cost benefits across both models. For Sonnet 3.7, journal tools with nonempty context achieve the strongest cost reduction at \$0.2490 (7.8\% reduction from baseline), while journal with empty context shows modest improvement at \$0.2651 (1.9\% reduction). Sonnet 4 exhibits similar patterns with journal nonempty context achieving \$0.2433 (9.0\% reduction) and journal empty context at \$0.2570 (3.9\% reduction). This pattern suggests that journal tools provide reliable benefits, with accumulated knowledge amplifying individual reflection by an additional 4-6\%.

Social media tools show mixed results with model-specific patterns. Sonnet 3.7 benefits modestly from social empty context (\$0.2639, 2.3\% reduction) but shows slight cost increases with nonempty context (\$0.2785, 3.1\% increase). In contrast, Sonnet 4 experiences significant cost increases with social tools, particularly social empty context (\$0.3293, 23.2\% increase). These divergent patterns indicate strong model compatibility effects, with Sonnet 3.7 adapting better to social coordination mechanisms than Sonnet 4.

The combined journal-social variants consistently increase costs across both models, ranging from 14.6\% to 52.1\% increases.This indicates that multiple similar overlapping tools may require additional differentiation or coordination to allow agents to utilize them effectively.

\subsubsection{Turn Efficiency}

\begin{table}[h]
  \caption{Average API Call Turns}
  \label{tab:turn-performance}
  \centering
  \begin{tabular}{lcc}
    \toprule
    Configuration & Sonnet 3.7 & Sonnet 4 \\
    \midrule
    Baseline & 42.20 & 40.96 \\
    Journal (Empty) & 41.39 (-1.9\%) & 43.52 (+6.3\%) \\
    Journal (Nonempty) & 43.40 (+2.8\%) & 42.41 (+3.5\%) \\
    Social (Empty) & 46.26 (+9.6\%) & 52.24 (+27.5\%) \\
    Social (Nonempty) & 46.79 (+10.9\%) & 49.13 (+19.9\%) \\
    Journal-Social (Empty) & 58.52 (+38.7\%) & 54.18 (+32.3\%) \\
    Journal-Social (Nonempty) & 50.17 (+18.9\%) & 54.73 (+33.6\%) \\
    \bottomrule
  \end{tabular}
\end{table}

Turn efficiency results show mixed patterns with generally modest changes from baseline. For Sonnet 3.7, only journal empty context achieves a meaningful reduction (41.39 vs 42.20 baseline, 1.9\% improvement), while other variants show increases ranging from 2.8\% to 38.7\%. Sonnet 4 demonstrates increases across all variants, with journal variants showing relatively modest increases (3.5-6.3\%) but social and combined variants requiring substantially more turns.

Unlike cost performance, turn efficiency shows minimal improvements, with most collaborative variants requiring additional API calls to perform a write, read, or search call. 

\subsubsection{Time Performance} 

\begin{table}[h] 
\caption{Average Duration (seconds)} 
\label{tab:time-performance} 
\centering 
\begin{tabular}{lcc} 
\toprule 
Configuration & Sonnet 3.7 & Sonnet 4 \\ 
\midrule 
Baseline & 94.9 & 99.7 \\ 
Journal (Empty) & 97.5 (+2.7\%) & 101.1 (+1.4\%) \\ 
Journal (Nonempty) & 88.3 (-7.0\%) & 97.2 (-2.5\%) \\ 
Social (Empty) & 90.3 (-4.9\%) & 117.5 (+17.8\%) \\ 
Social (Nonempty) & 94.7 (-0.2\%) & 120.2 (+20.5\%) \\ 
Journal-Social (Empty) & 142.0 (+49.5\%) & 133.8 (+34.2\%) \\ 
Journal-Social (Nonempty) & 110.2 (+16.1\%) & 126.4 (+26.8\%) \\ 
\bottomrule 
\end{tabular} 
\end{table} 

Time performance exhibits considerable variability with no consistent pattern of improvement. Sonnet 3.7 shows the best time reduction with journal nonempty context (88.3s vs 94.9s baseline, 7.0\% improvement), while other variants show mixed results. Sonnet 4 demonstrates a similar pattern: the journal-nonempty variant yields only a modest improvement (97.2s vs 99.7s, 2.5\% gain), whereas most social and combined variants require substantially more time.

Time results reinforce that social collaborative tools involve overhead costs that are only justified on sufficiently challenging problems. The mixed time performance suggests that tool benefits depend on problem difficulty -- easy problems suffer from unnecessary overhead while hard problems benefit from enhanced reasoning capability.

\subsubsection{Token Usage Analysis}

\begin{table}[h]
\caption{Average Token Usage - Full Dataset}
\label{tab:token-usage-full}
\centering
\begin{tabular}{lcccccc}
\toprule
Configuration & Model & Input & Cache Creation & Cache Read & Output & Total \\
\midrule
Baseline & Sonnet 3.7 & 77 & 13,812 & 369,291 & 5,552 & 388,732 \\
         & Sonnet 4 & 81 & 13,067 & 380,818 & 4,811 & 398,777 \\
\midrule
Journal (Empty) & Sonnet 3.7 & 75 & 12,475 & 401,985 & 5,267 & 419,802 \\
                & Sonnet 4 & 77 & 12,602 & 422,790 & 4,950 & 440,420 \\
\midrule
Journal (Nonempty) & Sonnet 3.7 & 76 & 12,022 & 374,932 & 5,016 & 392,045 \\
                    & Sonnet 4 & 77 & 11,983 & 401,290 & 4,780 & 418,130 \\
\midrule
Social (Empty) & Sonnet 3.7 & 74 & 12,850 & 402,390 & 4,986 & 420,300 \\
               & Sonnet 4 & 83 & 15,619 & 553,092 & 5,767 & 574,560 \\
\midrule
Social (Nonempty) & Sonnet 3.7 & 75 & 13,741 & 429,415 & 5,371 & 448,602 \\
                   & Sonnet 4 & 84 & 14,636 & 473,973 & 5,538 & 494,231 \\
\midrule
Journal-Social (Empty) & Sonnet 3.7 & 76 & 17,396 & 645,989 & 7,850 & 671,312 \\
                       & Sonnet 4 & 84 & 16,039 & 563,298 & 6,226 & 585,647 \\
\midrule
Journal-Social (Nonempty) & Sonnet 3.7 & 69 & 14,853 & 498,676 & 6,039 & 519,637 \\
                           & Sonnet 4 & 82 & 15,481 & 576,916 & 5,993 & 598,472 \\
\bottomrule
\end{tabular}
\end{table}

Token usage patterns across the full dataset reveal the mechanisms underlying the mixed performance effects observed in business metrics. Analysis of token allocation provides insights into how collaborative tools affect agent reasoning processes and resource consumption.

\textbf{Output Token Efficiency}: The most successful cost-reduction variants consistently generate fewer output tokens compared to baseline. Sonnet 3.7 journal nonempty produces 5,016 output tokens versus 5,552 baseline (-9.6\%), while Sonnet 4 journal nonempty generates 4,780 versus 4,811 baseline (-0.6\%). Given that output tokens cost \$15 per million versus \$3 per million for input tokens, these reductions in expensive output generation directly contribute to cost savings.

\textbf{Resource Allocation Patterns}: Successful variants demonstrate more efficient resource allocation rather than increased compute consumption. Journal tools with nonempty context show modest increases in total token usage (+0.9\% for Sonnet 3.7, +4.8\% for Sonnet 4) while achieving significant cost reductions, indicating better utilization of cheaper input and cache operations relative to expensive output generation.

\textbf{Model-Specific Resource Usage}: Token patterns explain the divergent performance between models. Sonnet 4 social (empty) shows dramatically increased cache reads (553,092 vs 380,818 baseline, +45.2\%) and higher output tokens, correlating with its 23.2\% cost increase. In contrast, successful Sonnet 3.7 variants demonstrate more balanced resource allocation.

\textbf{Tool Overhead Effects}: Combined journal-social variants consistently show the highest token consumption across all categories, with total usage increases ranging from 33-73\%. This pattern explains why combined tools often hurt performance—the overhead of managing multiple collaborative interfaces outweighs individual benefits when problems don't require extensive reasoning scaffolding.

These token usage patterns confirm that collaborative tools function as reasoning amplifiers rather than compute scaling mechanisms, with performance gains arising from more efficient resource allocation rather than increased token consumption.

\section{Test Completion Metrics}
\label{app:test-completion-analysis}

\begin{table}[h]
  \caption{Challenge Completion Rates (100\% passing tests)}
  \label{tab:completion-rates}
  \centering
  \begin{tabular}{lcc}
    \toprule
    Configuration & Sonnet 3.7 & Sonnet 4 \\
    \midrule
    Baseline & 99.0\% & 98.0\% \\
    Journal (Empty) & 100.0\% & 98.0\% \\
    Journal (Nonempty) & 99.0\% & 99.0\% \\
    Social (Empty) & 100.0\% & 95.1\% \\
    Social (Nonempty) & 98.0\% & 99.0\% \\
    Journal-Social (Empty) & 98.0\% & 98.0\% \\
    Journal-Social (Nonempty) & 98.0\% & 95.1\% \\
    \bottomrule
  \end{tabular}
\end{table}

\begin{table}[h]
  \caption{Overall Test Pass Rates (passed tests/total tests)}
  \label{tab:pass-rates}
  \centering
  \begin{tabular}{lcc}
    \toprule
    Configuration & Sonnet 3.7 & Sonnet 4 \\
    \midrule
    Baseline & 99.8\% & 99.8\% \\
    Journal (Empty) & 100.0\% & 99.4\% \\
    Journal (Nonempty) & 99.1\% & 99.3\% \\
    Social (Empty) & 100.0\% & 99.1\% \\
    Social (Nonempty) & 99.8\% & 99.8\% \\
    Journal-Social (Empty) & 99.3\% & 99.7\% \\
    Journal-Social (Nonempty) & 98.8\% & 97.3\% \\
    \bottomrule
  \end{tabular}
\end{table}

Test completion metrics demonstrate that collaborative tools maintain high performance standards without sacrificing solution quality. Both models achieve challenge completion rates above 95\% across all variants, with several tool variants achieving perfect or near-perfect completion rates. Overall test pass rates remain consistently above 97\%, indicating that the additional reasoning overhead from collaborative tools does not compromise challenge completions.

\section{Complete Search Interaction Example}
\label{app:complete-search}

This appendix presents the complete interaction sequence showing how an agent systematically leveraged collaborative tools for upfront knowledge gathering before implementing a bowling game scoring system.

\textbf{Initial Problem Approach:}
\begin{quote}
\small\textit{"Let me explore the bowling directory structure and check for previous journal entries or social media posts about bowling challenges."}
\end{quote}

\textbf{Journal Search Query:}
The agent searched journal entries with the query "bowling scoring kata challenge" (limit: 5 entries).

\textbf{Search Results Discovery:}
The search returned 5 relevant entries, with the top result (score: 0.524) showing a completed bowling challenge:
\begin{quote}
\small\textit{"Bowling Challenge - COMPLETED [CHECKMARK] Final implementation successfully passes all 31 tests including: Basic scoring (strikes, spares, regular frames), 10th frame special rules..."}
\end{quote}

\textbf{Knowledge Synthesis:}
\begin{quote}
\small\textit{"Excellent! I found a previous journal entry that shows this bowling challenge has been completed before. Let me read that entry to understand the solution approach, then check social media to see what teammates have shared."}
\end{quote}

\textbf{Detailed Journal Entry Review:}
The agent retrieved the complete journal entry (path: 3f45012d-f2f5-408e-bd71-2e16b62ef068), which contained detailed implementation insights including:
- Key architectural decisions (frame-based vs roll-based thinking)
- Common pitfalls (10th frame complexity, game completion detection)  
- Debugging strategies (running individual failing tests)
- Refactoring lessons (simpler implementations often work better)

\textbf{Social Media Search:}
The agent searched for posts tagged "bowling" (limit: 10), discovering three relevant posts documenting the journey from initial attempt through debugging to successful completion.

\textbf{Knowledge Integration:}
\begin{quote}
\small\textit{"Perfect! I have excellent context from both my journal and social media posts. Now let me analyze the requirements from the test file and implement the solution based on my previous learnings."}
\end{quote}

This interaction demonstrates sophisticated upfront research behavior where the agent systematically gathered, reviewed, and synthesized accumulated knowledge before beginning implementation—behavior that emerged organically without explicit instruction.

\section{Agent Tool Instructions}
\label{app:tool-instructions}

This appendix presents the complete instructions provided to agents for each collaborative tool configuration, demonstrating the minimal, non-prescriptive guidance that resulted in sophisticated emergent behaviors.

\subsection{Journal-Only Configuration Instructions}

\begin{quote}
\textbf{Coding with Your Journal}

You're working on coding challenges, and you have access to a personal journal where you can reflect on your work.

\textbf{Feel free to write in your journal whenever you want}

Check out what you've written before. Review problems you've worked on, discoveries you've made, or just browse through recent entries. Use it like you would any journal - when you want to reflect, need to organize your thoughts, or want to review your progress.

\textbf{Write entries when you feel like it}

Record whatever feels worth documenting:
\begin{itemize}
\item Something cool you figured out
\item A frustrating bug you're dealing with
\item A quick win or breakthrough
\item Just thoughts about what you're working on
\item Notes for future reference
\end{itemize}

\textbf{Search and browse your entries}
\begin{itemize}
\item Leverage the work your team has done before! You should check what work has already been done - your or your team's previous entries might save you from reinventing solutions. If you do search for relevant articles and then read the ones which seem relevant
\end{itemize}

The search tools will show you recent entries and let you semantically search as you like. Review your past work and see what patterns emerge in your problem-solving approach.

\textbf{No pressure}

This is meant to be natural and helpful. Write if you want to, browse when you feel like it, or ignore it entirely if you're in the zone. There's no requirement to use your journal in any particular way.

\textit{Focus on solving your coding challenges. The journal is just there if you want to use it.}
\end{quote}

\subsection{Social Media-Only Configuration Instructions}

\begin{quote}
\textbf{Coding with Social Media}

You're working on coding challenges, and you have access to a social media platform where you can connect with teammates.

\textbf{Feel free to browse social media whenever you want}

Check out what others are posting about. See what problems they're working on, what they're discovering, or just scroll through recent posts. Use it like you would any social media - when you want a break, need inspiration, or are curious what's happening.

\textbf{Post when you feel like it}

Share whatever feels worth sharing:
\begin{itemize}
\item Something cool you figured out
\item A frustrating bug you're dealing with
\item A quick win or breakthrough
\item Just thoughts about what you're working on
\item Encouragement for teammates
\end{itemize}

\textbf{Search and browse casually}

The search tools will show you recent posts and let you filter by tags. Don't expect perfect results - just browse around and see what catches your eye.

\textbf{No pressure}

This is meant to be natural and relaxed. Post if you want to, browse when you feel like it, or ignore it entirely if you're in the zone. There's no requirement to use social media in any particular way.

\textit{Focus on solving your coding challenges. The social media is just there if you want to use it.}
\end{quote}

\subsection{Combined Configuration Instructions}

\begin{quote}
\textbf{Coding with Your Journal and Social Media}

You're working on coding challenges, and you have access to both a personal journal and a social media platform where you can connect with teammates.

\textbf{Feel free to use either whenever you want}

Check out what you've written before in your journal or browse what others are posting on social media. Review problems you've worked on, discoveries you've made, or see what teammates are sharing. Use them like you would naturally - when you want to reflect, need inspiration, want to organize your thoughts, or are just curious what's happening.

\textbf{Write or post when you feel like it}

Record or share whatever feels worth documenting:
\begin{itemize}
\item Something cool you figured out
\item A frustrating bug you're dealing with
\item A quick win or breakthrough
\item Just thoughts about what you're working on
\item Notes for future reference
\item Encouragement for teammates
\end{itemize}

\textbf{Search and browse your entries and posts}
\begin{itemize}
\item Leverage the work your team has done before! You should check what work has already been done - your previous journal entries or your team's social media posts might save you from reinventing solutions. If you do search for relevant articles and then read the ones which seem relevant
\item The search tools will show you recent entries and posts, letting you semantically search through both your personal notes and team discussions
\item Review your past work and see what patterns emerge in your problem-solving approach
\item Browse casually through social media to see what catches your eye
\end{itemize}

\textbf{Journal vs Social Media}

Use your \textbf{journal} for:
\begin{itemize}
\item Personal reflection and deeper thoughts
\item Detailed technical notes
\item Private problem-solving process
\item Things you want to remember for yourself
\end{itemize}

Use \textbf{social media} for:
\begin{itemize}
\item Sharing wins and discoveries with the team
\item Getting input from teammates
\item Casual updates and encouragement
\item Building team connections
\end{itemize}

Or don't worry about the distinction and just use whatever feels right in the moment.

\textbf{No pressure}

This is meant to be natural and helpful. Write in your journal, post to social media, browse when you feel like it, or ignore both entirely if you're in the zone. There's no requirement to use either tool in any particular way.

\textit{Focus on solving your coding challenges. The journal and social media are just there if you want to use them.}
\end{quote}

\section{Hard Question Selection}
\label{app:hard-question-selection}

\subsection{Threshold Sensitivity Analysis}

To evaluate the robustness of our hard-questions definition, we examined performance at the $\mu + 1\sigma$ threshold, which represents problems requiring substantially more computational resources than the baseline distribution. This more stringent threshold identifies 4 problems for Sonnet 3.7 (bowling, connect, forth, react) and 2 problems for Sonnet 4 (transpose, two-bucket), representing 11.8\% and 5.9\% of the benchmark, respectively.

At this threshold, collaborative tools demonstrate even more dramatic performance improvements. Sonnet 3.7 achieves cost reductions ranging from 22.5\% to 45.7\% across most variants, with social (empty) delivering the strongest reduction (\$0.455 vs \$0.838 baseline, 45.7\% reduction) and journal (nonempty) achieving 33.8\% reduction (\$0.555 vs \$0.838). Turn efficiency improvements are similarly substantial, with journal-social nonempty requiring 38.5\% fewer API calls (56.5 vs 92.0 baseline) and social empty achieving 31.6\% reduction (62.9 turns).

Sonnet 4 shows strong selective benefits despite the small sample size, with journal nonempty delivering 63.9\% cost reduction (\$0.406 vs \$1.127 baseline) and 37.8\% duration improvement (155.4s vs 402.3s baseline). The journal variants consistently outperform baseline across all metrics, while social tools show mixed results with social nonempty achieving 23.6\% cost reduction but social empty increasing costs by 16.9\%.

However, the more restrictive threshold substantially reduces sample sizes to n=5--11 per configuration for Sonnet 3.7 and n=5--6 for Sonnet 4, compared to n=11--17 at the $\mu + 0.5\sigma$ threshold. While the effect sizes are larger and more dramatic, the reduced statistical power limits the reliability of these results for formal hypothesis testing. The consistency of improvement patterns across both thresholds provides confidence in the underlying mechanisms, but the $\mu + 0.5\sigma$ threshold offers a better balance between capturing genuinely challenging problems and maintaining adequate sample sizes for robust statistical analysis.

These results reinforce our core finding that collaborative tools provide the greatest benefits when agents face problems at the limits of their capabilities, with effect magnitude scaling inversely with problem frequency in the benchmark distribution.

\begin{table}[h]
  \centering
  \caption{Sonnet 3.7 Hard Questions - Cost Performance with Distribution ($\mu + 1\sigma$ threshold)}
  \begin{tabular}{lllrrrrr}
  \toprule
  Configuration & Context & n & Mean & Median & P90 & P95 & P99 \\
  \midrule
  Baseline & – & 11 & \$0.838 & \$0.761 & \$1.413 & \$1.541 & \$1.643 \\
  Social & Empty & 11 & \$0.455 (-45.7\%) & \$0.499 & \$0.638 & \$0.668 & \$0.693 \\
  Social & Nonempty & 11 & \$0.724 (-13.6\%) & \$0.560 & \$1.680 & \$2.001 & \$2.258 \\
  Journal & Empty & 11 & \$0.725 (-13.5\%) & \$0.541 & \$1.756 & \$1.917 & \$2.046 \\
  Journal & Nonempty & 11 & \$0.555 (-33.8\%) & \$0.438 & \$0.913 & \$1.001 & \$1.072 \\
  Journal-Social & Empty & 10 & \$0.591 (-29.5\%) & \$0.579 & \$1.067 & \$1.141 & \$1.201 \\
  Journal-Social & Nonempty & 11 & \$0.649 (-22.5\%) & \$0.499 & \$1.025 & \$1.318 & \$1.553 \\
  \bottomrule
  \end{tabular}
\end{table}
  
\begin{table}[h]
  \centering
  \caption{Sonnet 4 Hard Questions - Cost Performance with Distribution ($\mu + 1\sigma$ threshold)}
  \begin{tabular}{lllrrrrr}
  \toprule
  Configuration & Context & n & Mean & Median & P90 & P95 & P99 \\
  \midrule
  Baseline & – & 6 & \$1.127 & \$0.878 & \$2.038 & \$2.353 & \$2.605 \\
  Social & Empty & 6 & \$1.317 (+16.9\%) & \$1.025 & \$2.421 & \$2.892 & \$3.270 \\
  Social & Nonempty & 6 & \$0.861 (-23.6\%) & \$0.827 & \$1.322 & \$1.360 & \$1.390 \\
  Journal & Empty & 5 & \$0.672 (-40.4\%) & \$0.541 & \$1.092 & \$1.137 & \$1.174 \\
  Journal & Nonempty & 5 & \$0.406 (-63.9\%) & \$0.342 & \$0.573 & \$0.635 & \$0.685 \\
  Journal-Social & Empty & 6 & \$1.042 (-7.5\%) & \$1.014 & \$1.716 & \$1.766 & \$1.806 \\
  Journal-Social & Nonempty & 6 & \$1.081 (-4.1\%) & \$1.007 & \$1.995 & \$2.112 & \$2.206 \\
  \bottomrule
  \end{tabular}
\end{table}

\section{Infrastructure Issues and Dataset Completion}
\label{nonrandom-docker-errors}

\subsection{Docker Configuration Failures}
  
  During initial experimental runs, we identified a Docker container configuration issue affecting 2.5\% of challenge attempts (approximately 35 out of 1,428 total runs). The issue occurred when unit test libraries attempted memory cleanup after test timeouts, causing container failures for challenges that lacked specific Python testing libraries. These failures were non-random and infrastructure-related rather than model performance issues.
  
  The failure pattern included:
  \begin{itemize}
  \item 10 pairs where both empty and nonempty runs failed
  \item 4 baseline configuration failures  
  \item 2 cases where empty runs failed but second pass completed
  \item 6 cases where empty runs passed but nonempty runs failed
  \end{itemize}
  
  \subsection{Conservative Remediation Methodology}
  
  To complete the dataset while preserving experimental integrity, we implemented a conservative approach prioritizing data quality over potential performance gains:
  
  \textbf{Double Failures (Both Empty and Non-Empty):} Runs were executed on isolated team IDs, eliminating any shared context but ensuring clean experimental conditions.
  
  \textbf{Empty Run Failures:} Only the failed empty run was re-executed on a new team ID, allowing the nonempty run to proceed with whatever limited context existed.
  
  \textbf{Non-Empty Run Failures:} The original empty run data was preserved, and only the nonempty run was re-executed using the established team ID, maintaining full experimental context.
  
  \textbf{Minimal Social Configuration Impact:} Of the 10 double-failure pairs requiring isolated re-runs, 7 involved social variants (0.49\% of total dataset). Given that social nonempty variants consistently showed the weakest performance across both models, any potential information advantage from re-running these specific cases would bias results toward variants that were already underperforming, making our reported effects conservative estimates.
  
  \textbf{Potential Social Tool Effects:} The remediation process may have inadvertently benefited some social nonempty variants by providing cleaner information environments. Of the 10 double-failure re-runs requiring isolated team IDs, 7 involved social variants, potentially reducing the accumulated "noise" that makes tag-based filtering challenging. This could partially explain the unexpectedly stable performance of Sonnet 4's social nonempty variant through extreme percentiles.

\subsection{Robustness Validation}

  To validate the stability of our findings, we compared results across datasets before and after infrastructure remediation:
  
  \textbf{Effect Consistency:} Comparing results before and after infrastructure remediation shows stable performance patterns with surgical changes only where remediation occurred.
  
  \textbf{Sonnet 3.7 Hard Questions:} Social empty remained completely unchanged (\$0.436, 39.4\% cost reduction), demonstrating that unaffected configurations were unaltered by remediation. Remediated variants showed modest shifts: social nonempty changed from 37.8\% to 21.5\% cost reduction, and journal-social nonempty from 24.4\% to 15.2\% reduction. Journal empty showed larger changes from 41.5\% to 15.5\% cost reduction, reflecting infrastructure fixes in configurations that experienced failures.
  
  \textbf{Sonnet 4 Hard Questions:} The remediation process affected problem composition, with the dataset changing from 5 to 4 hard questions (removing zebra-puzzle), while baseline costs shifted from \$0.777 to \$0.805. Despite these changes, core collaborative tool patterns remained consistent: journal nonempty maintained strong performance (41.3\% to 40.0\% cost reduction) and journal empty showed sustained benefits (31.5\% to 30.9\% reduction). Configuration rankings remained unchanged with journal tools consistently outperforming other variants.
  
  \textbf{Validation of Conservative Approach:} The stability of unaffected variants (such as Sonnet 3.7 social empty showing identical performance) alongside targeted shifts in remediated configurations confirms that infrastructure fixes addressed specific failures without introducing systematic bias. The preservation of relative performance rankings across both models demonstrates that core collaborative mechanisms remained intact.
  
  The consistency of collaborative tool benefits—with effect patterns preserved despite infrastructure remediation and minor compositional changes—demonstrates that our findings reflect genuine performance mechanisms rather than artifacts of specific experimental configurations.
  
  \subsection{Statistical Implications}
  
  The infrastructure remediation did not systematically bias results toward any particular configuration. The conservative approach ensures that reported improvements represent lower bounds on collaborative tool effectiveness, as any information leakage would inflate rather than deflate performance benefits.
  
  Full dataset metrics remained stable throughout remediation (within $\pm$0.02 cost variation), confirming that infrastructure issues affected only a small subset of runs without systematically altering the overall experimental conclusions.


\begin{thebibliography}{99}
  \bibitem{anthropic2024mcp}
  Anthropic. (2024, November 25). Introducing the model context protocol. \textit{Anthropic News}. \url{https://www.anthropic.com/news/model-context-protocol}
  
  \bibitem{chen2023agentverse}
  Chen, W., Su, Y., Zuo, J., Yang, C., Yuan, C., Chan, C.-M., Yu, H., Lu, Y., Hung, Y.-H., Qian, C., Qin, Y., Cong, X., Xie, R., Liu, Z., Sun, M., \& Zhou, J. (2023). AgentVerse: Facilitating multi-agent collaboration and exploring emergent behaviors. \textit{arXiv preprint arXiv:2308.10848}. \url{https://doi.org/10.48550/arXiv.2308.10848}
  
  \bibitem{espinosa2001shared}
  Espinosa, J. A., Kraut, R. E., Lerch, J. F., Slaughter, S. A., Herbsleb, J. D., \& Mockus, A. (2001). Shared mental models and coordination in large-scale, distributed software development. In *Proceedings of the 22nd International Conference on Information Systems (ICIS 2001)* (pp. 513–518). Association for Information Systems. https://aisel.aisnet.org/icis2001/64/

  \bibitem{gibson1979ecological}
  Gibson, J. J. (1979). \textit{The ecological approach to visual perception}. Houghton Mifflin.
  
  \bibitem{kiyokawa2023verbalization}
  Kiyokawa, S., Uchida, N., \& Liu, M. (2023). Verbalization toward others facilitates insight problem solving. In M. Goldwater, F. K. Anggoro, B. K. Hayes, \& D. C. Ong (Eds.), Proceedings of the 45th Annual Conference of the Cognitive Science Society (pp. 3166–3171). Cognitive Science Society. https://escholarship.org/uc/item/4312q9wk

  \bibitem{norman2013doet}
  Norman, D. (2013). \textit{The design of everyday things: Revised and expanded edition}. Basic Books.
  
  \bibitem{park2023generative}
  Park, J. S., O'Brien, J., Cai, C. J., Morris, M. R., Liang, P., \& Bernstein, M. S. (2023). Generative agents: Interactive simulacra of human behavior. In *Proceedings of the 36th Annual ACM Symposium on User Interface Software and Technology* (UIST '23) (pp. 1–22). ACM. https://doi.org/10.1145/3586183.3606763

  \bibitem{shinn2023reflexion}
  Shinn, N., Cassano, F., Berman, E., Gopinath, A., Narasimhan, K., \& Yao, S. (2023). Reflexion: Language agents with verbal reinforcement learning. \textit{arXiv preprint arXiv:2303.11366}. \url{https://doi.org/10.48550/arXiv.2303.11366}

  \bibitem{wei2022chain}
  Wei, J., Wang, X., Schuurmans, D., Bosma, M., Ichter, B., Xia, F., Chi, E., Le, Q. V., \& Zhou, D. (2022). Chain-of-thought prompting elicits reasoning in large language models. In Advances in Neural Information Processing Systems (Vol. 35, pp. 24824–24837). Curran Associates, Inc. https://arxiv.org/abs/2201.11903

  \bibitem{wu2024autogen}
  Wu, Q., Bansal, G., Zhang, J., Wu, Y., Li, B., Zhu, E., Jiang, L., Zhang, X., Zhang, S., Liu, J., Wang, Y., \& Li, M. (2024). AutoGen: Enabling next-gen LLM applications via multi-agent conversation. In Proceedings of the International Conference on Learning Representations (ICLR 2024). https://doi.org/10.48550/arXiv.2308.08155

  \bibitem{yu2014understanding}
  Yu, X., \& Petter, S. (2014). Understanding agile software development practices using shared mental models theory. \textit{Information and Software Technology}, 56(8), 911--921. \url{https://doi.org/10.1016/j.infsof.2014.02.010}
  
  \bibitem{yao2023react}
  Yao, S., Zhao, J., Yu, D., Du, N., Shafran, I., Narasimhan, K., \& Cao, Y. (2023). ReAct: Synergizing reasoning and acting in language models. In \textit{Proceedings of the International Conference on Learning Representations (ICLR 2023)}. \url{https://openreview.net/forum?id=WE_vluYUL-X}
  
  \bibitem{zhao2025survey}
  Zhao, W. X., Zhou, K., Li, J., Tang, T., Wang, X., Hou, Y., Min, Y., Zhang, B., Zhang, J., Dong, Z., Du, Y., Yang, C., Chen, Y., Chen, Z., Jiang, J., Ren, R., Li, Y., Tang, X., Liu, Z., Liu, P., Nie, J.-Y., \& Wen, J.-R. (2025). A survey of large language models. \textit{arXiv preprint arXiv:2303.18223}. \url{https://doi.org/10.48550/arXiv.2303.18223}
  
  \end{thebibliography}
\end{document}